\documentclass[letterpaper, 10pt, conference]{ieeeconf}      

\IEEEoverridecommandlockouts                              

\usepackage{amsmath} 
\usepackage{amssymb}  
\usepackage{graphicx}
\usepackage{subcaption}
\usepackage{float}
\usepackage{amsmath}
\usepackage{subcaption}
\usepackage{overpic}
\usepackage{algorithm}
\usepackage{algpseudocode}
\usepackage{multirow} 
\usepackage{soul}

\usepackage{pifont}
\usepackage{xcolor}
\usepackage{soul} 
\usepackage[colorlinks,
linkcolor=blue,
 anchorcolor=blue,
 citecolor=blue]{hyperref}
 \usepackage{overpic}
 \usepackage[font=small,labelfont=bf]{caption}
 
\newcommand{\action}{\mathbf{a}}
\newcommand{\state}{\mathbf{s}}
\newcommand{\cmdpenv}{\mathcal{M}}
\newcommand{\risk}{\epsilon_\text{risk}}
\newcommand{\safetythreshold}{\epsilon_\text{safe}}
\newcommand{\taskpolicy}{\pi_\text{task}}
\newcommand{\recoverypolicy}{\pi_\text{recovery}}
\newcommand{\safetycritic}{Q_\text{risk}}
\newcommand{\gammasafe}{\gamma_\text{safe}}
\newcommand{\gammatask}{\gamma_\text{task}}
\newcommand{\stiffness}{\mathbf{K}}
\newcommand{\damping}{\mathbf{D}}
\newcommand{\wrenchee}{\mathbf{w}^\text{EE}}
\newcommand{\posvec}{\mathbf{p}}

\newcommand{\removed}[1]{{\color{gray} }}

\title{
\textbf{Bresa}: \textbf{B}io-inspired \textbf{Re}flexive \textbf{Sa}fe Reinforcement Learning \\for Contact-Rich Robotic Tasks
}

\author{Heng Zhang$^{1,2,*}$, Gokhan Solak$^{1,*}$, Arash Ajoudani$^{1}$
\thanks{* These two authors contribute equally to the work.}
\thanks{
This work was supported by the Horizon Europe Project TORNADO (GA 101189557).} 
\thanks{$^{1}$ Human-Robot Interfaces and Interaction Lab, Istituto Italiano di Tecnologia, Genoa, Italy.\newline
        e-mails: \mbox{{\{heng.zhang,gokhan.solak,arash.ajoudani\}@iit.it}}}%
\thanks{$^{2}$~Ph.D. program of national interest in Robotics and Intelligent Machines (DRIM) and Università di Genova, Genoa, Italy.}
}

\begin{document}


\maketitle

\begin{abstract}
Ensuring safety in reinforcement learning (RL)-based robotic systems is a critical challenge, especially in contact-rich tasks within unstructured environments. While the state-of-the-art safe RL approaches mitigate risks through safe exploration or high-level recovery mechanisms, they often overlook low-level execution safety,  where reflexive responses to potential hazards are crucial. Similarly, variable impedance control (VIC) enhances safety by adjusting the robot's mechanical response , yet lacks a systematic way to adapt parameters, such as stiffness and damping throughout the task. In this paper, we propose Bresa, a \textbf{B}io-inspired \textbf{Re}flexive Hierarchical \textbf{Sa}fe RL method inspired by biological reflexes. Our method decouples task learning from safety learning, incorporating a \textit{safety critic} network that evaluates action risks and operates at a higher frequency than the task solver. Unlike existing recovery-based methods, our \textit{safety critic} functions at a low-level control layer, allowing real-time intervention when unsafe conditions arise. The task-solving RL policy, running at a lower frequency, focuses on high-level planning (decision-making), while the \textit{safety critic} ensures instantaneous safety corrections.
We validate Bresa on multiple tasks including a contact-rich robotic task, demonstrating its reflexive ability to enhance safety, and adaptability in unforeseen dynamic environments. Our results show that Bresa outperforms the baseline, providing a robust and reflexive safety mechanism that bridges the gap between high-level planning and low-level execution.
Real-world experiments and supplementary material are available at project website \href{https://jack-sherman01.github.io/Bresa/}{https://jack-sherman01.github.io/Bresa}.

\end{abstract}

\section{INTRODUCTION} \label{sec:intro}

\begin{figure}[bt]
    \centering
    \begin{tabular}{cc}
        \includegraphics[width=0.60\linewidth]{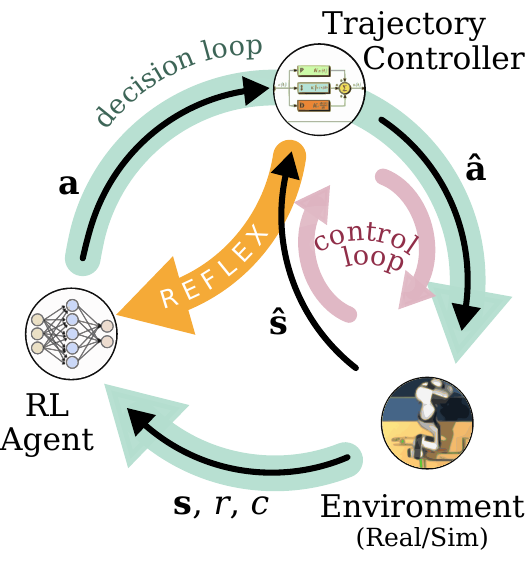} 
        &  
        \includegraphics[width=0.20\linewidth]{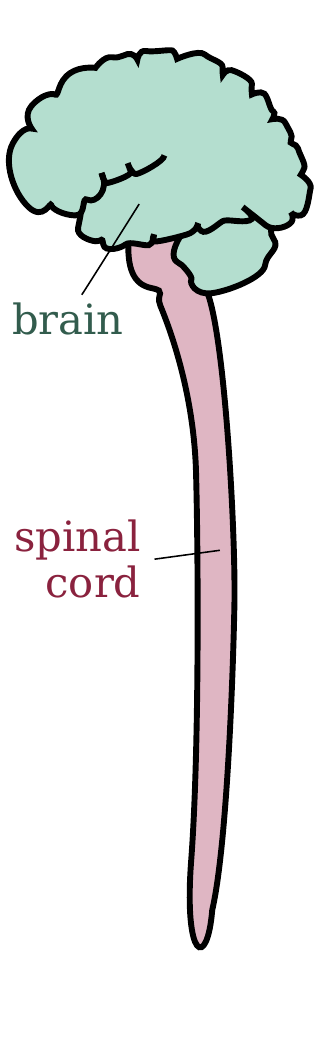} 
         \\
        (a) & (b) 
    \end{tabular}
    \caption{\textbf{a)} Bresa framework. 
    The RL agent operates at the \textit{decision loop}, planning the high-level action $\action$ that is executed by the trajectory controller. 
    The controller operates at a high-frequency \textit{control loop}, executing the low-level action $\hat{\action}$ based on the state feedback $\hat{\state}$ at each control step.
    The reflex mechanism gives the system a quick reaction capability by interrupting the control loop in the case of high risk.
    \textbf{b)} A simplified illustration of the human central nervous system. While high-level decisions are made in the brain, safety-related reflexes are managed by the spinal cord, allowing for faster responses that override slower, more complex decision-making processes.}
    \label{fig:intro}
\end{figure}


Robotic actions in the real world present two major challenges: the complexity of unstructured environments and the safety hazards associated with physical interactions~\cite{suomalainen2022survey}. RL-based robotic systems have the potential to address both challenges to enable effective automated learning and exploration in such environments~\cite{gu2024safesurvey}. Traditionally, the complexity challenge has received significant attention, while the safety challenge has gained focus more recently, especially in contact-rich tasks~\cite{suomalainen2022survey}. Drawing inspiration from the animal kingdom, where evolutionary processes have led to the development of solutions to these challenges, we propose a novel approach. Specifically, this paper draws on the reflex mechanisms inherent to vertebrates to enhance the safety and robustness of RL systems.

The complexity challenge can be mitigated by imposing a hierarchy between long-term task-level actions and short-term motor-level actions. Early RL studies have shown the advantage of decomposing complex tasks into smaller subtasks \cite{dietterich2000hierarchical}. 
Furthermore, it was recently shown that considering actions in task-space leads to more efficient learning, in comparison to using joint-space actions \cite{martin2019variable}. 
Thus, we assume a hierarchy between the high-level decision-making and low-level control loops as shown in Fig.~\ref{fig:intro}.a. 

The safety challenge has gained more attention recently due to the black-box nature of traditional learning-based RL models in safety-critical environments, under the umbrella of safe RL \cite{brunke2022safe}. Specifically, we follow the hierarchical safe RL methods, which answer the task-solving and safety problems through separate learned components. 
A crucial component is a \textit{safety critic} network that estimates the risk of failure to limit access to risky states \cite{bharadhwaj2020conservative}. 
Furthermore, a separate \textit{recovery policy} can be deployed in addition to the usual \textit{task policy} for learning more complex risk-evasive behavior \cite{thananjeyan2021recovery}. 
Integrating variable impedance control (VIC) into hierarchical safe RL is shown to enhance contact-rich interaction safety further through adaptable compliant behavior \cite{HengSRL-VIC}. 
However, existing safety methods primarily function at the decision-making stage, often overlooking risks that emerge during low-level action execution. 



Addressing safety in the decision-making loop leaves the system vulnerable to many safety-critical events occurring at the control stage. 
An action may be initially evaluated as safe, however, the risk may increase during execution because of the dynamicity, stochasticity and partial observability of the environment~\cite{nguyen2021robust,noseworthy2025forge}. 
Specifically, sensor noise, dynamical effects, or delays in actuation can lead to deviations from the intended trajectory, pushing the system into unsafe states. External factors such as changing environmental conditions or unforeseen obstacles can further compromise safety during execution. 
Furthermore, contact-rich tasks are particularly prone to control-time risks, as forming and breaking physical contact introduces highly non-linear and discontinuous dynamics, requiring similarly dynamic responses~\cite{ajoudani2018progress,kuo2021uncertainty}. 
Therefore, integrating a dedicated low-level safety mechanism is essential for ensuring instantaneous corrective actions during execution.


We propose the \textbf{B}iomimetic \textbf{Re}flexive Hierarchical \textbf{Sa}fe RL (\textbf{Bresa}) to address this gap. 
Our method is inspired by the reflex mechanism common to all vertebrates. As illustrated for humans in Fig.~\ref{fig:intro}.b, reflexes are managed by the spinal cord, bypassing the complex reasoning in the brain~\cite{bear2020neuroscience}. 
Similarly, we place the safety evaluation in the high-frequency control loop to immediately interrupt the execution when the risk exceeds a threshold, as shown in Fig.~\ref{fig:intro}.a. 
The reflex triggers the \textit{recovery policy} that aims to escape the danger, instead of the \textit{task policy} that aims to solve the task while it is safe.

We evaluate our method on both a 2-dimensional navigation task and a contact-rich maze exploration task. For the maze exploration task we first train our method in Mujoco simulator, and then transfer to a real-world robot platform. We compare Bresa to the baseline hierarchical safe RL method \cite{HengSRL-VIC} where the safety-checking and task-solving happens together in the \textit{decision loop}. 
The results show that Bresa significantly decreases safety violations, and consequently improves efficiency in task learning. 
The novel method ensures a reflexive and adaptive response to potential hazards, significantly improving safety in contact-rich and uncertain environments.

An important aspect of our method is that it is designed to work on unknown contact-rich environments. 
Bresa learns the safety notion from data and reactively avoids risks.
Our framework allows contacts with the environment, that is fundamental in contact-rich tasks.
In that regard, it differs from the works like \cite{fan2024learn,liu2023safe} that model the safety constraints geometrically and project the RL action into a safe tangential space to avoid obstacles. 
Another related work \cite{bing2023safety}, uses an RL agent for planning and an MPC for safe low-level execution. This work also defines the constraints as contact avoidance. The MPC ensures safety at control-level, given the obstacle positions and a dynamics model.

The RL-based contact-rich applications usually take advantage of adaptable impedance/admittance controllers \cite{zhang2023efficient,aflakian2024robust}. 
These two works rely on simulators for safe training, and then transfer the learned model to real-world through techniques such as domain randomization. 
However, differently than our method, they do not explicitly answer the safety problem.
\cite{zhu2022contact} proposes a control-level solution by monitoring the external force and counteracting it through null-space control. 
However, this approach assumes the expected forces produced by impedance controller are already safe. 
In contrast, our method monitors the state and impedance action continuously to ensure safety.
In summary, the main contributions of this work are as follows:
\begin{enumerate}
    \item We propose a hierarchical safe RL approach inspired by animal reflexes, where the safety mechanism operates at a higher frequency than the task solver to ensure rapid responses in contact-rich environments.
    \item Unlike existing recovery-based methods that focus on high-level planning, our method introduces a \textit{safety critic} at the low-level control layer, enabling reflexive real-time intervention and improving execution safety in unforeseen situations. 
    \item We demonstrate that our approach improves safety and learning efficiency on multiple tasks: a contact-rich robotic task, and a 2-dimensional navigation task.
\end{enumerate}

\removed{
\section{Related work}\label{sec:related-works} 

\subsection{Variable impedance control for contact-rich tasks}
Variable impedance control (VIC) allows a robot to adjust its compliance, thereby enhancing adaptability in contact-rich tasks. However, setting proper impedance parameters can be challenging when tasks or task dynamics vary. 
To address this problem, recent advancements in VIC have focused on integrating it with various learning strategies to improve performance across complex scenarios.
Zhou et al. \cite{zhou2025variable} explored an imitation learning-based VIC approach for collaborative mobile manipulators, highlighting the impact of variable impedance on task execution in industrial settings.
Zhang et al. \cite{zhang2023efficient} focused on learning variable impedance control for contact-rich manipulation tasks, leveraging inverse reinforcement learning to tackle force-related challenges.
\cite{okada2024contact} introduced contact models for improving variable impedance control through diffusion learning, further enhancing performance in manipulation tasks involving soft and deformable objects.
Furthermore, advancements in optimizing impedance parameters for dynamic manipulation tasks have been studied by several works, such as \cite{tsuji2024adaptive,li2024learning,hejrati2023nonlinear} that focus on enhancing the adaptability and robustness of VIC in dynamic and uncertain environments.
These efforts demonstrate the continued evolution of VIC in response to the growing complexity of contact-rich robotic tasks.

\subsection{Safe RL for contact-rich tasks}
Safe RL methods for contact-rich tasks focus on extending conventional RL methods by incorporating constraint mechanisms that ensure physical and operational safety during interactions. 
Recent studies have introduced innovative frameworks combining RL with safety mechanisms to improve both task performance and the robot's interaction with the environment.
Rastegarpanah et al.\cite{aflakian2024robust} presented a curriculum-based domain randomization approach that enhances RL-based learning for robust and safe contact-rich task performance, particularly in path-following tasks.
Zhang et al.\cite{zhang2023efficient} introduced a sim-to-real transfer approach for safe and stable RL in contact-rich tasks, particularly highlighting its utility in manipulation and insertion tasks where safety is crucial.
Furthermore, Sleiman et al. \cite{sleiman2024guided} presented a safe RL framework that integrates impedance control for robust multi-contact loco-manipulation, enhancing safety in real-time learning by predicting action safety before execution.
Shi et al. \cite{fan2024learn} explored safety considerations in manipulation tasks, providing solutions for ensuring safe interaction during dynamic robotic tasks and mitigating risk during operation.
In some real-world robotic tasks, continuous physical contact is inevitable. SRL-VIC~\cite{HengSRL-VIC} is a model capable of performing tasks in a continuous physical contact-rich environment without violating safety thresholds, employing both safe RL and variable stiffness modulation. 
These approaches not only focus on ensuring safety but also on improving the efficiency of learning by leveraging model-free RL techniques in real-world applications.

In contrast to these works, our method focuses on enhancing safety in contact-rich tasks by introducing a bio-inspired reflex mechanism with safe RL method that ensures rapid responses to potential hazards during execution.



\subsection{safety guarantee in low-level control for contact-rich tasks}
Ensuring safety in high-frequency controllers (often required in real-time robotic systems) is crucial because the robot's decisions must be executed quickly and accurately without compromising safety. Among the state of the art methods, those that provide safety guarantees in high-frequency control tend to rely heavily on model-based methods because of their ability to predict and control the system's dynamics in real-time. 
\cite{bing2023safety} proposed a safety mechanism in low-level control for contact-rich tasks. The safety mechanism ensures instantaneous safety corrections when unsafe conditions arise. The task-solving RL policy, running at a lower frequency, focuses on high-level planning (decision-making), while the impedance controller runs in a high frequency. The method is validated on a contact-rich robotic task, demonstrating its ability to enhance safety reflexively and adaptability in dynamic environments.
In~\cite{HengSRL-VIC}, an impedance controller is integrated with the safe RL framework, allowing adaptive stiffness regulation based on safety evaluations, thus enhancing compliance and robustness in dynamic, unstructured environments during interaction.
\cite{zhu2022contact} proposed a contact-rich task learning method that combines RL with a contact model to ensure safe and efficient task execution.
\cite{liu2023safe} introduced a framework for safe exploration in reinforcement learning applied to robotic tasks. It enforces safety constraints by operating in the tangent space of the constraint manifold, ensuring that learned policies avoid collisions and adhere to safety requirements during high-frequency control operations.

However, the majority of the works are model-based, as they provide more structured, predictable, and provably safe policies. In contrast, we propose a hierarchical approach that combines model-free learning with safety constraints, achieving a balance between flexibility and safety. This allows the robot to adapt to unforeseen situations while maintaining safe operation.
}

\begin{figure*}[tb]
    \centering
    \begin{tabular}{ccc} 
         \includegraphics[width=0.14\linewidth]{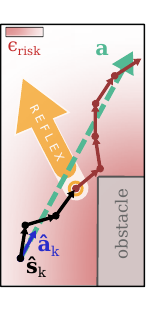}
        \hspace{-4mm}
        \vspace{-1mm}
         &
         \includegraphics[width=0.57\linewidth]{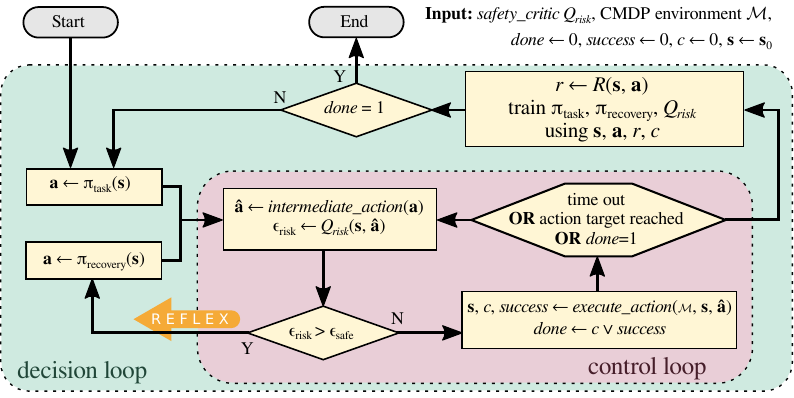} 
         &
         \includegraphics[width=0.24\linewidth]{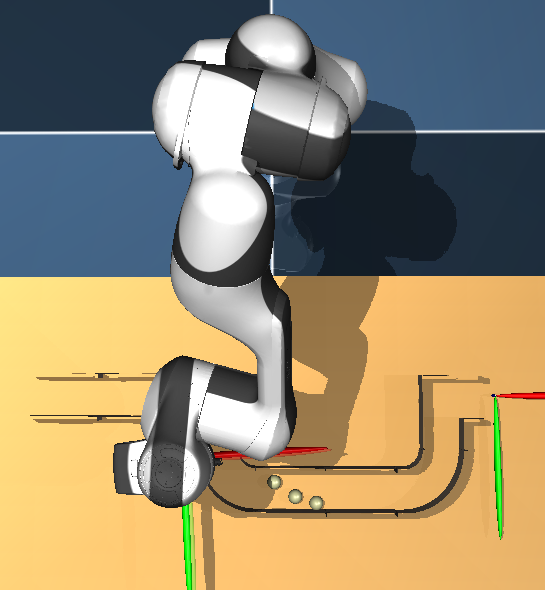} 
         \\
         (a) & (b) & (c)
    \end{tabular}
    \caption{
    \textbf{a)} Reflex mechanism on an obstacle avoidance scenario. Even when the high-level state-action pair $(\state, \action)$ is evaluated to be safe, an intermediate state-action pair $(\hat{\state},\hat{\action})$ may entail high risk ($\risk > \epsilon_\text{safe}$) and trigger the reflex mechanism. 
    The stochasticity of the environment leads to a drift in the outcomes of minor actions $\hat{\action}$.
    \textbf{b)} Flowchart of the Bresa algorithm. 
    We color-coded the \textit{decision loop}, \textit{control loop} and \textit{reflex} for comparison to Fig.~\ref{fig:intro}.a. 
    We reuse $\state$ instead of showing $\hat{\state}$ to simplify the structure, however, they are equivalent in the control loop.
    \textbf{c)} Maze exploration environment in the Mujoco simulator. The robot physically interacts with the maze walls and the obstacles through an end-effector flange equipped with F/T sensor.  
    }\label{fig:method}
\end{figure*}

\section{Bio-inspired Hierarchical Reflexive Safe RL}\label{sec:methodology}
The main concept of Bresa methodology is summarized in Fig.~\ref{fig:intro}. A key element of our approach is the bio-inspired reflex mechanism, designed to rapidly interrupt the control loop and swiftly respond to predicted dangers.
Our method follows the existing line of hierarchical safe RL works that separate the task learning and safety learning by learning a value function for assessing the risk of taking an action in a given state \cite{bharadhwaj2020conservative,thananjeyan2021recovery,HengSRL-VIC}. 

We advance the state-of-the-art (as our baseline controller described in Sec.~\ref{sec:hierarchical-safe-rl}) by establishing a bio-inspired hierarchy between the system's task-solving and safety-ensuring components. In Sec.~\ref{sec:reflex-method}, we describe the details of the novel reflex mechanism and how we answer the design challenges related to it.

The low-level trajectory controller in our framework (Fig.~\ref{fig:intro}.a) is tasked to execute the high-level action $\action$ in the control loop. It is abstracted to be any controller that achieves subtasks of $\action$. In this work, we use linear interpolation between the state $\state$ at the beginning of the action and the desired target state. In the contact-rich maze exploration task, we employ a VIC to compliantly execute the obtained linear trajectory. The controllers are detailed in Sec.~\ref{sec:traj-control}.  In the following we begin with the safe RL problem definition.


\subsection{Safe RL Problem}
We follow the constrained Markov decision process (CMDP) formulation \cite{altman1995constrained} to define the safe RL problem. A CMDP environment $\cmdpenv{=} (\mathcal{S}, \mathcal{A}, R, P, \gammatask, \mu, \mathcal{C})$ consists of  the state space $\mathcal{S}$, the action space $\mathcal{A}$, reward function $R: \mathcal{S} \times \mathcal{A} \rightarrow \mathbb{R}$, the state transition probability $P$, reward discount factor $\gammatask \in(0,1)$, the starting state distribution $\mu$ and the safety constraints $\mathcal{C}=\left\{\left(c_i: \mathcal{S} \rightarrow\{0,1\}, \chi_i \in \mathbb{R}\right) \right\}$, where $c_i{=}1$ indicates the violation of $i^{th}$ constraint.
$\cmdpenv$ is implemented differently for each environment, and we present these in Sec.~\ref{sec:nav-task} and \ref{sec:maze-task}.

\subsection{Hierarchical Safe RL} \label{sec:hierarchical-safe-rl}

Here, we describe our baseline method to solve the safe RL problem. Please refer to \cite{thananjeyan2021recovery} for a more detailed formalization of this method.  
We train two policies online, $\taskpolicy$ for completing the task and $\recoverypolicy$ for evading unsafe states. 
We train also a safety critic $\safetycritic$ to estimate risk incurred by a state-action pair $(\state \in \mathcal{S}, \action \in \mathcal{A})$.
Given a safety threshold $\safetythreshold \in \mathbb{R}$, the estimated risk $\risk \in \mathbb{R}$ of the task action determines which policy is sampled in the decision loop:
\begin{equation}
\begin{aligned}\label{eq:rrl-action}
    \risk &= \safetycritic(\state, \taskpolicy(\state)),\\
    \action &= \begin{cases}
                \recoverypolicy(\state) & \text{if } \risk > \safetythreshold\\
                \taskpolicy(\state) & \text{otherwise}.
             \end{cases}
\end{aligned}
\end{equation}
$\safetycritic$ is pre-trained before the task exploration phase using simple action samples that are procedurally collected offline. 
Details of the pre-training are discussed later in Sec.~\ref{sec:sim-exp}. The training of $\safetycritic$ continues online during the RL exploration. 
We use a different discount factor ($\gammasafe$) in critic training as the risk depends on shorter term effects than the task. 

In reference to Fig.~\ref{fig:intro}.a, the action sampling \eqref{eq:rrl-action} happens in the RL agent node, while the \textit{reflex} does not happen in the baseline method.

\begin{algorithm}[t]
\caption{Bresa}
\label{alg:bresa}
\begin{algorithmic}[1]  
    \State \textbf{Input:} CMDP environment $\cmdpenv$, safety critic $\safetycritic$
    \State \textbf{Output:} Policies $\pi_\text{task}, \pi_\text{recovery},$ safety critic $\safetycritic$
    \State $\textit{reflex} \gets 0, \textit{done} \gets 0, \textit{success} \gets 0, c \gets 0, \state \gets \state_0$
    \Repeat \Comment{decision loop}
        \State $\action \gets \begin{cases}
                \recoverypolicy(\state) & \text{if } \textit{reflex} = 1\\
                \taskpolicy(\state) & \text{otherwise}.
             \end{cases}$
        \State $\textit{reflex} \gets 0$
        \Repeat \Comment{control loop}
            \State $\hat{\action} \gets $\textit{intermediate\_action}$(\action)$
            \State $\risk \gets \safetycritic(\state, \hat{\action})$
            \If{$\risk > \safetythreshold$}
                \State $\textit{reflex} \gets 1$
                \State \textbf{break} control loop \Comment{reflex}
            \EndIf
            \State $\state, c, \textit{success} \gets $\textit{execute\_action}$(\cmdpenv, \state, \hat{\action})$ \Comment{$\hat{\state} = \state$}
            \State $\textit{done} \gets c \lor \textit{success}$
        \Until{$\text{time out} \lor \text{action target reached} \lor \textit{done} = 1$}
        \State $r \gets R(\state, \action)$
        \State train $\pi_\text{task}, \pi_\text{recovery}, \safetycritic$ using $\state,\action, r, c$
    \Until{$\textit{done} = 1$}
\end{algorithmic}
\end{algorithm}

\subsection{Reflex Mechanism} \label{sec:reflex-method}
In this section, we give the complete flowchart (Fig.~\ref{fig:method}.b) and pseudo-code (Alg.~\ref{alg:bresa}) of the Bresa algorithm. 
The flowchart showcases the explicit relationships between the \textit{decision} and \textit{control loops}.
The task-solving policy operates at the low-frequency \textit{decision loop}, while the safety critic operates at the high-frequency \textit{control loop}. In case a danger arises, the safety critic has the power to interrupt the current controller command quickly and activate the recovery policy.
Please note that $\safetycritic$ is learned as a neural network that has predictive capabilities. Differently than a simple force check, it learns a complex multi-dimensional relationship that enables force contacts subject to other conditions. 

We illustrate the benefit of the reflex mechanism on an obstacle avoidance scenario in Fig.~\ref{fig:method}.a.
Given a high-level action $\action$, the low-level controller executes smaller intermediate actions that we call \textit{minor action} $\hat{\action}$. 
As the intermediate state $\hat{\state}$ drifts towards high-risk area due to the action noise, Bresa triggers a reflex to establish safety. 
We can observe a similar behaviour in our results later in Sec.~\ref{sec:nav-task}.

When the reflex happens, the intended action is interrupted, and consequently only a proportion of $\action$ gets executed. 
Training the models using this action decreases the transparency, and consequently undermines the learning performance. 
Thus, we add the executed action $\action'$, rather than the intended action $\action$ into the training dataset.
Our preliminary studies has shown better performance with this approach.
For position-based actions, we define the executed actions as $\action_k' = \posvec_{k+1}-\posvec_k$, where $\posvec_k$ is the position of the agent before the $k^{th}$ action $\action_k$.

From the statistics point of view, running the safety critic in the control loop creates a bias in its input towards smaller actions. 
When we bootstrap the $\safetycritic$ with uniformly sampled actions as in \cite{HengSRL-VIC}, it cannot learn a good initial critic and the learning performance decreases significantly. 
For this reason, we also implement a bias in the offline data collection procedure, giving higher probability to smaller actions. We call it as \textit{minorization} and describe how we achieve this for each task in Sec.~\ref{sec:nav-task} and \ref{sec:maze-task}.

\subsection{Trajectory Controller}\label{sec:traj-control}

In Fig.~\ref{fig:method}.b, we abstract the roles of the trajectory controller and the environment as the function handles \textit{intermediate\_action} and \textit{execute\_action}.
The implementation of these depend on the task.

The \textit{intermediate\_action} function returns the next control action $\hat{\action}$ towards the high-level action $\action$ goal. 
In this work, we use linear interpolation between the current position and the action target position to determine the trajectory points. 
This function should keep an internal memory of the trajectory and iteration index. 

The \textit{execute\_action} function encapsulates the low-level control rule, execution of the motor commands, environment dynamics, and observation of the outcomes, including the updated state $\state$, constraint violation $c$ and task success. 

In our robotic application, we apply Cartesian impedance control with variable stiffness and damping \{$\stiffness, \damping\}\in \mathbb{R}^{6\times6}$, assuming quasi-static conditions. The damping matrix $\damping$ is formed proportionally to $\stiffness$ as described in \cite{ott2008cartesian}. 
We calculate the desired end-effector wrench as  
\begin{equation}
\wrenchee =  \stiffness\Tilde{\boldsymbol{x}} + \damping\dot{\Tilde{\boldsymbol{x}}},
    \label{eq:cartesian_impedance}
\end{equation}
where $\Tilde{\boldsymbol{x}}\in\mathbb{R}^{6}$ is the Cartesian pose error between the current pose in $\hat{\state}$ and the target pose implied by the action $\hat{\action}$. Accordingly, $\dot{\Tilde{\boldsymbol{x}}}$ is the velocity error between the desired and actual end-effector's velocity. The stiffness matrix is defined in the world frame. 

We run the control until either the target ($\|\Tilde{\boldsymbol{x}}\|<2$ \textit{mm}) or time limit is reached. 
The latter is needed when the target is behind a wall, and thus it can never be reached.


\section{Evaluation}
In the following evaluation experiments, we investigate whether the proposed algorithm: 1) enhances safety by reducing violations and 2) improves efficiency in terms of success-to-violation ratio. To assess these aspects, we compare our approach to the baseline method (Sec.~\ref{sec:hierarchical-safe-rl}) in two complex tasks. Firstly, we validate our concept in a complex 2D navigation task described in~\ref{subsubsec:nav}.  We initially used an easy-to-simulate navigation task to test our method with a large number of repetitions, because a contact-rich robotic task in a 3-dimensional physics simulation requires long computation times, making it hard to evaluate it iteratively under different conditions. Secondly, integrated with a 7-DoF robot, we apply this method to a contact-rich maze exploration task. We conduct both simulation (Sec.~\ref{subsubsec:maze}) and real-world experiments (Sec.~\ref{sec:real_world_exp}) with the maze task.

    

\subsection{Experiment setup}\label{sec:exp-setup} 
We designed two tasks in simulation: 2-D Navigation (OpenAI Gym) and Maze exploration tasks (
MuJoCo version 2.3.3), and deployed the maze exploration task in the real world. 


\subsubsection{Navigation Task}\label{subsubsec:nav}
A navigation task inspired by similar tasks in~\cite{thananjeyan2021recovery}, with more obstacles to increase the difficulty. The 2-D environment is depicted in Fig.~\ref{fig:offline_data}. In this task, the agent navigates on a 2-D plane to go from the start point (green) to the goal point (yellow) without touching the rectangular obstacles (blue). 
The task is in a rectangular area of 100 x 80 units, where 
six rectangular obstacles are placed in the task area forming multiple tight bottlenecks to overcome.

\subsubsection{Maze Exploration Task}\label{subsubsec:maze}
Adopted from~\cite{HengSRL-VIC}, the simulated maze exploration task (Fig~\ref{fig:method}.c) is a 3-D robotic contact-rich task, where the robot does not have access to vision, thus it has to complete the task using force feedback. 
We use a peg-shaped flange, 30 mm in diameter and 55 mm long, mounted on the robot's end-effector. The maze channel features four turns, measuring 50 mm in width and 70.35 cm in total length. To increase the dynamic aspect of the environment, we place three movable spheres inside the maze.
For the real-world setup please refer to~\ref{sec:real_world_exp}.


\subsection{Simulation experiments} \label{sec:sim-exp}

\begin{figure}[tb]
    \centering
    \begin{tabular}{cc}
        \begin{minipage}{0.45\linewidth}
            \centering
            \includegraphics[width=\linewidth]{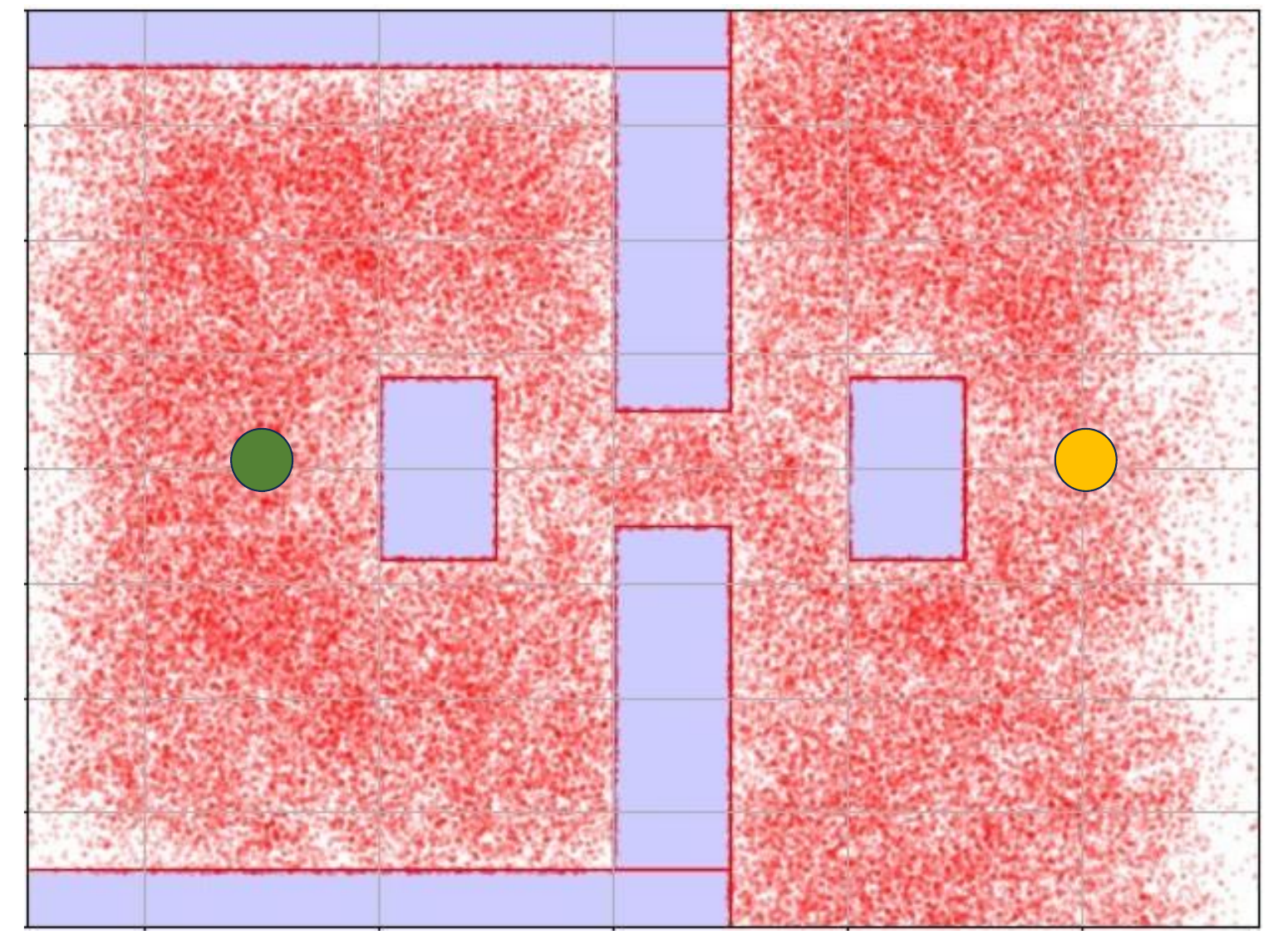}
        \end{minipage} &
        \begin{minipage}{0.47\linewidth}
            \centering
            \includegraphics[width=\linewidth]{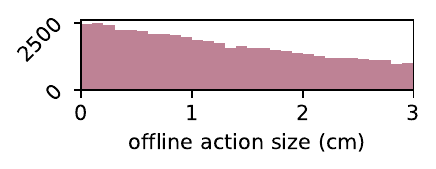}\\
            \includegraphics[width=\linewidth]{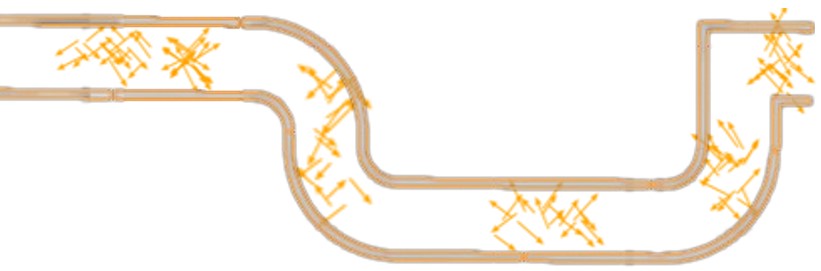}
        \end{minipage}
    \end{tabular}
    \caption{Offline data collection locations in both tasks. Left: navigation task. The green and yellow circles indicate start and goal points, and red dots indicate the sampled start positions. Upper right: histogram of exponentially sampled action sizes in the maze exploration task. Lower right: sampled action locations on the maze.}
    \label{fig:offline_data}
\end{figure}

\subsubsection{Navigation Task} \label{sec:nav-task}
In this task, the state space consists of the position $\posvec \in \mathbb{R}^2$ of the agent. The action is a change of position $\Delta\posvec \in \mathbb{R}^2$ in the range of $[-a_\text{max}, a_\text{max}]$. The constraint is violated ($c=1$) when $\posvec$ is inside an obstacle. The reward $R$ is based solely on the negative Euclidean distance to the goal.

We execute an action in a control loop through multiple \textit{minor actions} $\hat{\action}_k = m\action /\|\action\|$, for a minor action size $m$.  
When calculating the next state as the action outcome, we add a Gaussian noise perturbation $\action_\text{noise}$ as follows: 
    \begin{equation}
        \hat{\state}_{k+1} = \hat{\state}_k + \hat{\action}_k + \action_\text{noise},\quad \action_\text{noise} \sim \mathcal{N}(0, \sigma).
    \end{equation}
This helps us simulate the stochasticity of realistic tasks. In our main experiments $a_\text{max}{=}3, m{=}0.2, \sigma{=}0.02$.

(a). \textbf{Offline data collection}.
We randomly sample 100K offline data in the task area as shown in Fig.~\ref{fig:offline_data}, where red dots indicate starting points and light blue areas indicate obstacles. 
At each starting point, we sample up to 10 consecutive actions, terminated in case of a constraint.  The offline data consist of 100K tuples of [$\state$, $\action$, $\state'$, $c$]. 
In total, there are 5000 violations. To increase the possibility of violation, we ensure the percentage of constraint samples in all transitions not less than $5\%$ by discarding the safe transitions until reaching the desired percentage of violations. Otherwise, the data does not contain sufficient positive examples to learn the safety concept. 

We select offline action sizes through uniform sampling $\|\Delta\posvec\| \sim \mathcal{U}(-3, 3)$, however, for \textit{minorization}, we scale the selected action to \textit{minor action} size with $25\%$ chance.  

(b). \textbf{Training in simulation}
We train both the baseline and Bresa policies. We run 12 random seeds for each experiment to increase statistical accuracy. 
For each seed, we collect the offline data and pretrain the recovery policy and safety critic before the online training.
The agent has a horizon of $H{=}500$ steps to reach the goal in each episode. The episodes will be terminated immediately on constraint violation, i.e., collision with an obstacle. We identified different optimal discount factor values for each method: For the baseline $\gammasafe{=}0.80, \gammatask{=}0.94$; for Bresa $\gammasafe{=}0.65, \gammatask{=}0.95$. Please see Sec.~\ref{sec:ablation} for the details of the parameter study. The safety threshold is common for both methods $\safetythreshold{=}0.30$,
and the SAC temperature parameter $\alpha$ is auto-tuned.

\begin{figure}
    \centering
    \begin{subfigure}[b]{0.92\linewidth}
        \centering
        \includegraphics[width=\linewidth]{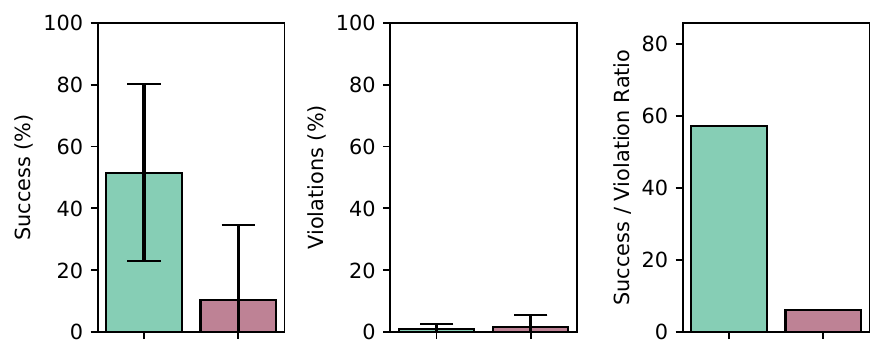}
        \caption{Performance in navigation task}
    \end{subfigure}
    \\
    \begin{subfigure}[b]{0.92\linewidth}
        \centering
        \includegraphics[width=\linewidth]{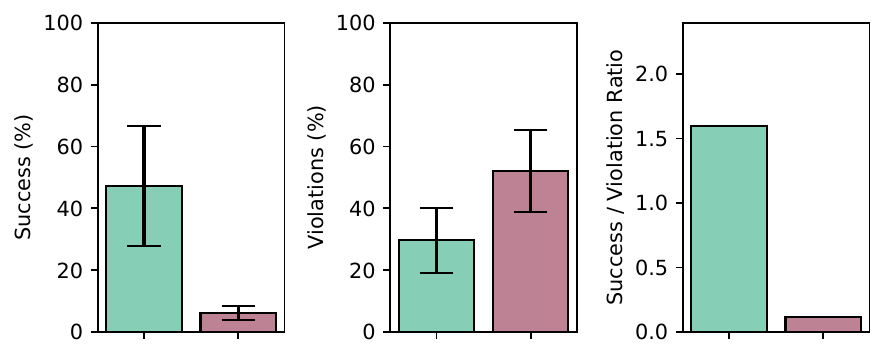}
        \caption{Performance in maze exploration task}
    \end{subfigure}
    \caption{Overall performance of Bresa in terms of success, violation, and the ratio between these two.}
    \label{fig:result_nav_maze}
\end{figure}

(c). \textbf{Results}.
Fig.~\ref{fig:result_nav_maze}.a demonstrates the overall performance of the proposed method in terms of task successes, constraint violations and success-violation ratio. Specifically, in total of 120 episodes training, Bresa achieves  61.75 in success and 1.08 in violation on average, while the baseline achieves only 12.50 in success and 2.08 in violation. Briefly, our method outperforms the baseline significantly in terms of the success-violation ratio. We also share our results with different hyperparameters and task parameters in Sec.~\ref{sec:ablation} to demonstrate the variability of the results and the consistency of the performance improvement.

\begin{figure}
    \centering
    \begin{overpic}[width=1\linewidth]{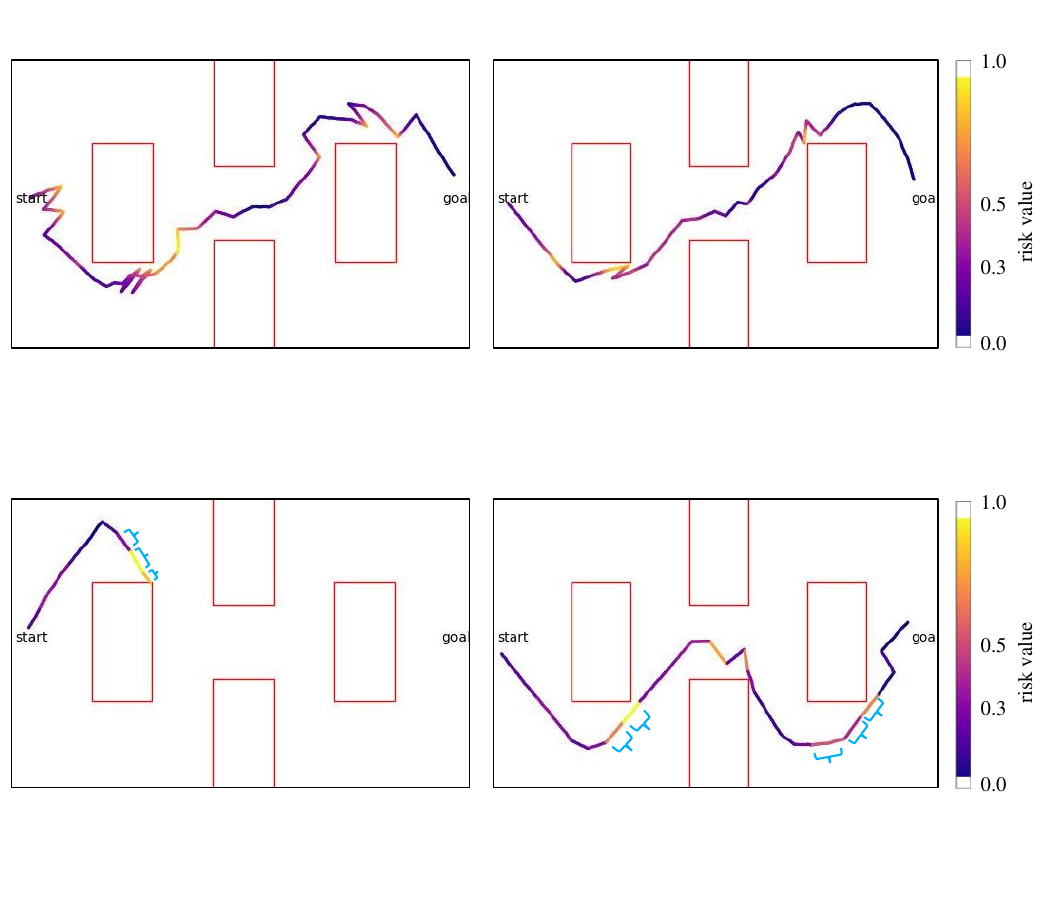}
    \put(38, 46){\small \bfseries (a) Bresa (Ours)}
    \put(38, 5){\small \bfseries (b) Baseline}
    \end{overpic}
    \caption{Reflexive mechanism in Navigation task during training. The risk value along with exploration is plotted in a colormap showing fine-grained risk prediction in our method while it is coarse-grained in the baseline. Blue annotations show single high-level actions. Note that the figures only show part of task space.}
    \label{fig:nav_traj}
\end{figure}

For a better insight on how the reflex mechanism works, we plot the trajectories of early and late training episodes with risk value colormaps for both methods in Fig.\ref{fig:nav_traj}. The Bresa trajectories show quick reflexive motions near the obstacles implying a more agile behaviour.

\subsubsection{Maze Exploration Task} \label{sec:maze-task}

In this task, the 12-dimensional state-space consists of the position, linear velocity, measured force and measured torque of the end-effector. 
The 4-dimensional action-space is formulated as the end-effector position change $\Delta\posvec$ in the range of $[-a_\text{max}, a_\text{max}]$ and stiffness  $K_1$, $K_2$ in the range of $[300, 1000]$. We use $a_\text{max}{=}0.03$ \textit{m} in this study. 
The safety constraint $c$ in this task is the contact force threshold that the robot cannot exceed when interacting with the environment. We set the force threshold as 30 N as a trade-off between safety and task success. We use the same reward function as in \cite{HengSRL-VIC}.

The stiffness actions $K_1$ and $K_2$ represent the stiffness along the major and minor axes of motion. We form the translational stiffness matrix in the world frame as $\mathbf{K}_t = \mathbf{R}_p^\top \mathbf{K}_a \mathbf{R}_p$, where 
\begin{equation*}
        \mathbf{R}_p {=}  \begin{bmatrix}
        \Delta p_x & -\Delta p_y  & 0\\
        \Delta p_y & \Delta p_x  & 0\\
    0 & 0  & 1\\
    \end{bmatrix}, \stiffness_a {=} \begin{bmatrix}
        K_1 & 0  & 0\\
        0 & K_2  & 0\\
    0 & 0  & K_z\\
    \end{bmatrix}.
\end{equation*}

The effective stiffness matrix is formed as $\stiffness = diag(\stiffness_t,\stiffness_r)$ with the fixed rotational stiffness $\stiffness_r=diag(100, 100, 0)$. Translational stiffness $K_z$ is fixed at $750$ N/m. 
$\stiffness$ is used in the VIC control as described in Sec.~\ref{sec:traj-control}.

(a). \textbf{Offline data collection}. Randomly sampled short-term action data was collected in five areas in the maze as shown in Fig.~\ref{fig:offline_data}, where orange arrows indicate random actions with uniformly random stiffness $K_1,K_2 \sim \mathcal{U}(300,1000)$ (N/m) and uniformly random direction $ \angle{\Delta\posvec} \sim \mathcal{U}(-\pi,\pi)$ (rad). 
The magnitude of the position change is sampled from an exponential distribution to favor smaller actions for \textit{minorization} as $\|\Delta\posvec\| \sim exp(\lambda)a_\text{max}$. 
We used $\lambda{=}1$ in our experiments and discarded values larger than $a_\text{max}$ (Fig.~\ref{fig:offline_data}). The offline data consist of 50000 tuples of [$\state$, $\action$, $\state'$, $c$] with 6034 constraint violations.

(b). \textbf{Training in simulation}
We train the policy in a contact-rich maze task, the basic structure of the network is similar to \cite{HengSRL-VIC}, SAC~\cite{SAC} is used to train task policy while DDPG~\cite{lillicrap2015continuous} is used for recovery policy and safety critic model. We run each experiment for 500 episodes and repeat the training for 12 seeds. We use the optimal discount factor parameters for both methods as $\gammasafe{=}0.675$ and $\gammatask{=}0.90$. The other important parameters are SAC temperature $\alpha{=}0.2$, safety threshold $\safetythreshold{=}0.45$, and horizon $H{=}300$.

\begin{figure}[tb]
    \centering
    \includegraphics[trim=0.2cm 0.2cm 0.2cm 0.1cm,width=0.97\linewidth]{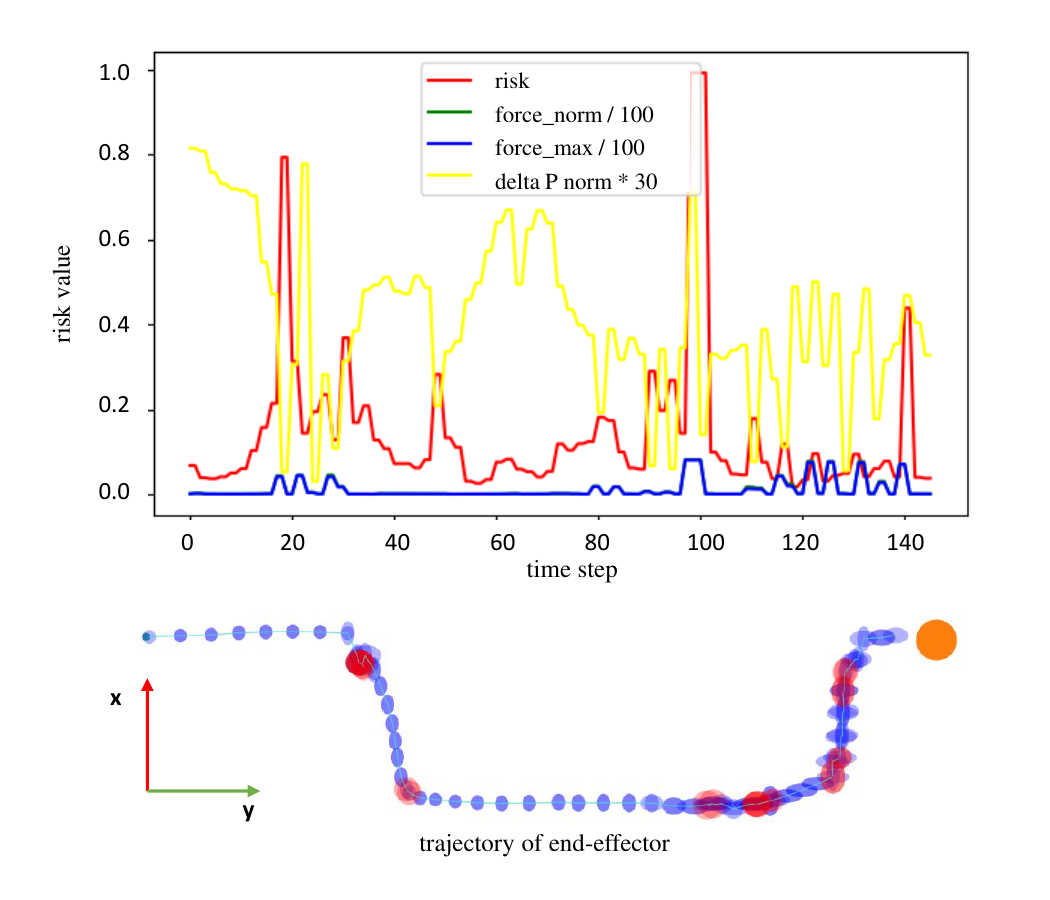}
    \caption{Illustration model performance from a specific episode. upper: reflexive response of \textit{risk critic} according to the current state and action. lower: trajectory of end-effector where the ellipsoids indicate stiffness and red color shows high-risk value.}
    \label{fig:risk_traj}
\end{figure}

(c). \textbf{Results}.
On the average of 12 seeds, Bresa achieves 81.6 success and 165.6 violation (ratio: 0.49) while the baseline achieves 10.5 success and 271.1 violation (ratio: 0.04), showing a significant improvement with the help of the reflexes.
Furthermore, we also report the results of best 3 seeds: Among 500 episodes of training, Bresa achieves 236.3 success and 148.0 violations on average, while the baseline has only 30.7 in success but 260.7 in violation on average. The best 3 seed results are shown in Fig.\ref{fig:result_nav_maze}.b, demonstrating that our method outperforms the baseline in terms of success, violation and the ratio between them. 

Moreover, we investigate how the reflex mechanism works in Bresa during training, showing the risk prediction in Fig.~\ref{fig:risk_traj}. We can see that the \textit{safety critic} predicts the risk value according to the current state and the next action.
The trajectory of the end-effector is shown in the lower part of Fig.~\ref{fig:risk_traj}, where the ellipsoids indicate stiffness and the red color shows the \textit{recovery action} in high-risk situations.
The reflex appears in the key location where a significant change occurs, such as the turns and obstacle contact.

\begin{figure}
    \centering
    \begin{subfigure}[b]{0.8\linewidth}
        \includegraphics[width=\linewidth]{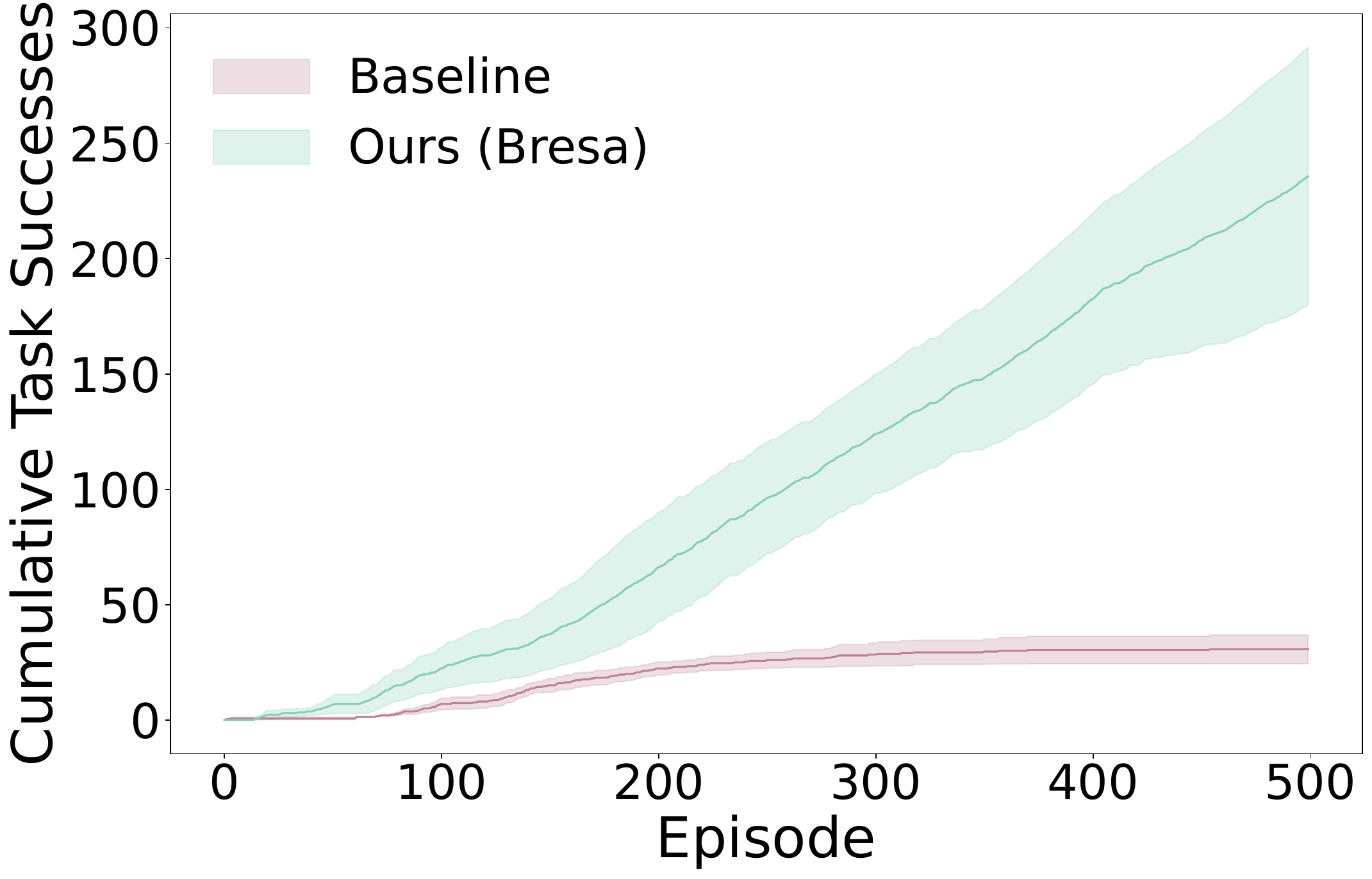}
        \caption{successes}
    \end{subfigure}


    \begin{subfigure}[b]{0.8\linewidth}
        \includegraphics[width=\linewidth]{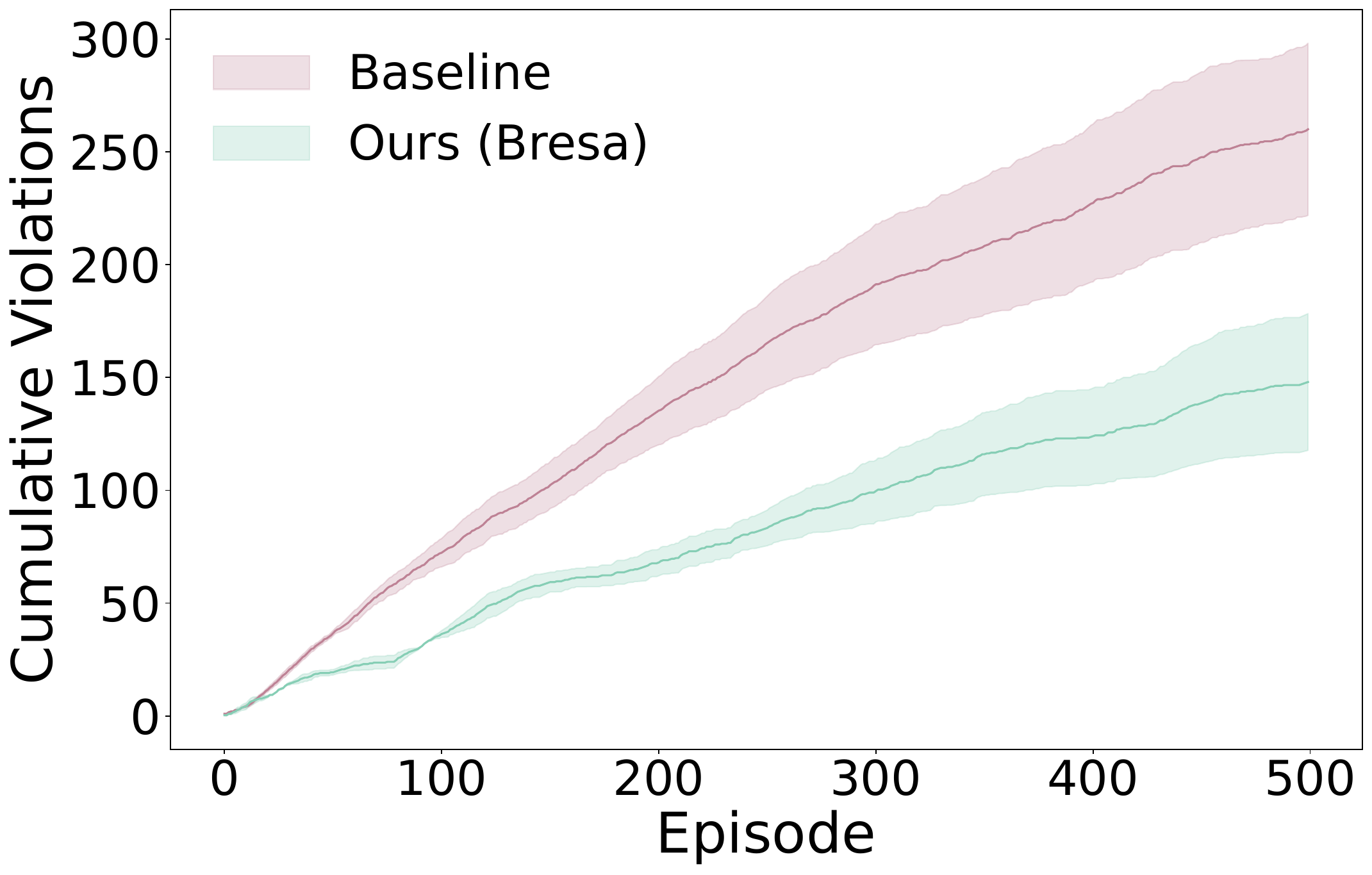}
        \caption{violations}
    \end{subfigure}
    \caption{Overall performance of Bresa in contact-rich maze exploration task in terms of cumulative successes and violations.}
    \label{fig:learning_curve}
\end{figure}

Fig.\ref{fig:learning_curve} shows the detailed learning process in terms of success and violation among 500 episodes. From the learning curve, it is clear that Bresa outperforms the baseline not only from the perspective of safety, but it also learns faster.

\subsection{Model Test}
We take the trained model of the best performing seed and run it 200 times to test the model performance in the contact-rich maze exploration task. The results are shown in Table~\ref{tab:model_test_sim}. The model achieves 98\% success rate and only 2 violations.
For further testing in real-world, please refer to~\ref{sec:real_world_exp}.

\begin{table}[tbp]
\centering
\caption{Model performance in maze exploration task}
\label{tab:model_performance}
\begin{subtable}[t]{0.9\linewidth}
    \centering
    \caption{Model test in simulation}
    \label{tab:model_test_sim}
    \begin{tabular}{c|c|c|c|c}
        \hline\hline
        Total runs & Success & Violation & Ratio & Success rate \\
        \hline
        200 & 196 & 2 & 98 & \textbf{98\%} \\
        \hline\hline
    \end{tabular}
\end{subtable}

\vspace{2mm}

\begin{subtable}[t]{0.9\linewidth}
    \centering
    \caption{Performance in real world}
    \label{tab:real_world}
    \begin{tabular}{c|c|c|c|c}
        \hline\hline
        Total runs & Success & Violation & Ratio & Success rate \\
        \hline
        10 & 10 & 0 & -- & \textbf{100\%} \\
        \hline\hline
    \end{tabular}
\end{subtable}
\end{table}

\begin{figure}
    \centering
    \begin{tabular}{cc}
        \begin{overpic}[width=0.51\linewidth]{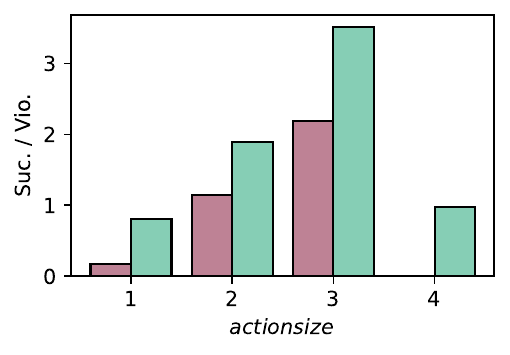}
            \put(18,58){\textbf{(a)}}
        \end{overpic}
        &
        \begin{overpic}[width=0.37\linewidth]{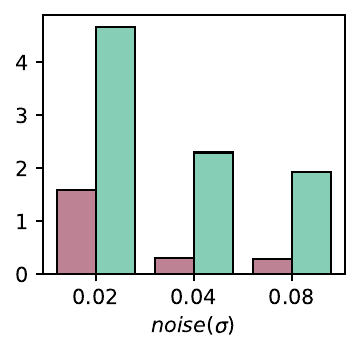}
            \put(75,80){\textbf{(b)}}
        \end{overpic} \\
        \multicolumn{2}{c}{
            \begin{overpic}[width=0.92\linewidth]{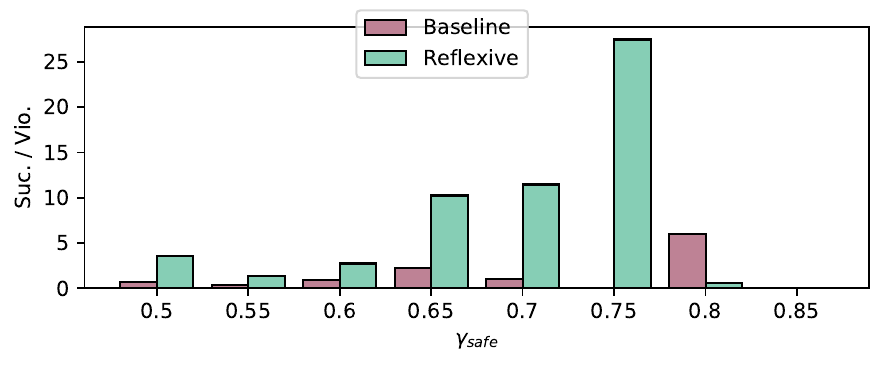}
                \put(12,34){\textbf{(c)}}
            \end{overpic}
        } \\
        \multicolumn{2}{c}{
            \begin{overpic}[width=0.93\linewidth]{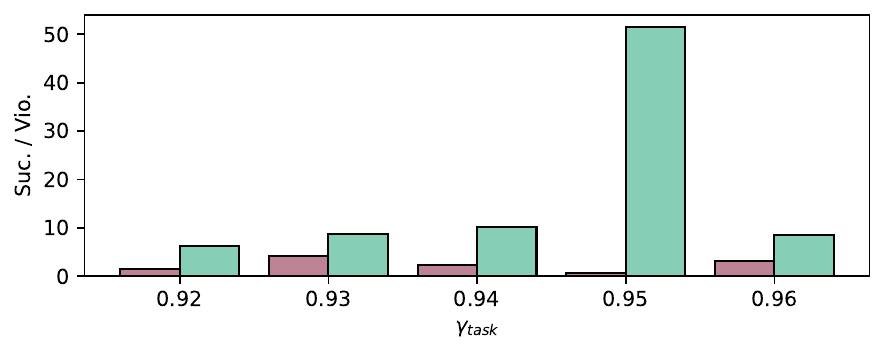}
                \put(12,34){\textbf{(d)}}
            \end{overpic}
        }
    \end{tabular}
    \caption{Success-violation ratio results over navigation task training with different \textit{action size} $a_\text{max}$, \textit{noise} $\sigma$, $\gammasafe$ and $\gammatask$ parameters shown in subfigure (a), (b), (c) and (d) respectively. }
    \label{fig:nav-results-bars}
\end{figure}

\subsection{Parameter studies} \label{sec:ablation} 
We performed parameter studies on the navigation task to evaluate the impact of different action size $a_\text{max}$, discount factors $\gammatask$, and noise scaler $\sigma$, for a comprehensive assessment of our method's performance. These experiments are essential for gaining a deeper understanding of the method's behavior, showcasing its advantages, and pinpointing opportunities for refinement. By conducting this analysis, we seek to underscore the versatility of our approach, validating its effectiveness and broad applicability across a wide range of safe RL techniques for robotic tasks involving extensive contact interactions.

\begin{enumerate}
    \item \textbf{Action size $a_\text{max}$}: We investigate the impact of different action sizes on the task performance. We compare the results of different $a_\text{max}$ values ($1, 2, 3, 4$). The results show that the action size of 3 achieves the best performance, as shown in Fig.~\ref{fig:nav-results-bars}.a. 
    As discussed in the introduction, high-level actions increase the sample efficiency by decreasing the decision-making frequency. 
    However, larger actions also has higher risk to violate constraints.

    \item  \textbf{Action noise $\sigma$:} 
    By default we use a noise scaler of $\sigma{=}0.02$, however, we also tested the effect of different $\sigma$ values ($0.02, 0.04, 0.08$). The results in Fig.~\ref{fig:nav-results-bars}.b show that our method is more robust to higher noise values.
    Comparatively, our method improved the success-violation ratio over the baseline by $292\%$ for $\sigma{=}0.02$, $750\%$ for $\sigma{=}0.04$, and $656\%$ for $\sigma{=}0.08$, in average over $20$ seeds.
    
    \item \textbf{Discount factors $\gammasafe,\gammatask$}: In our preliminary results we noticed the big influence of $\gammasafe, \gammatask$ parameters. Thus, we experimented with varying values of these. Each experiment is run 120 episodes with $24$ seeds. We used $\gammatask{=}0.94$ in $\gammasafe$ experiments, and $\gammasafe{=}0.65$ in $\gammatask$ experiments for both methods. 
    As seen in the results of Fig.~\ref{fig:nav-results-bars}.c and Fig.~\ref{fig:nav-results-bars}.d, Bresa obtains better success-violation trade-off consistently over different parameters, adding to the statistical confidence of our results. Both methods maintain a similar number of task successes during the training, however, our method decreases the number of constraint violations significantly. 
    The best $\gammasafe$ ($0.75$) is smaller than the best $\gammatask$ ($0.95$), because the safety is more immediate (short-term) than the exploration task.
    
\end{enumerate} 


\begin{figure}
    \centering
    \includegraphics[trim=1.2cm 1cm 1cm 1cm,width=0.97\linewidth]{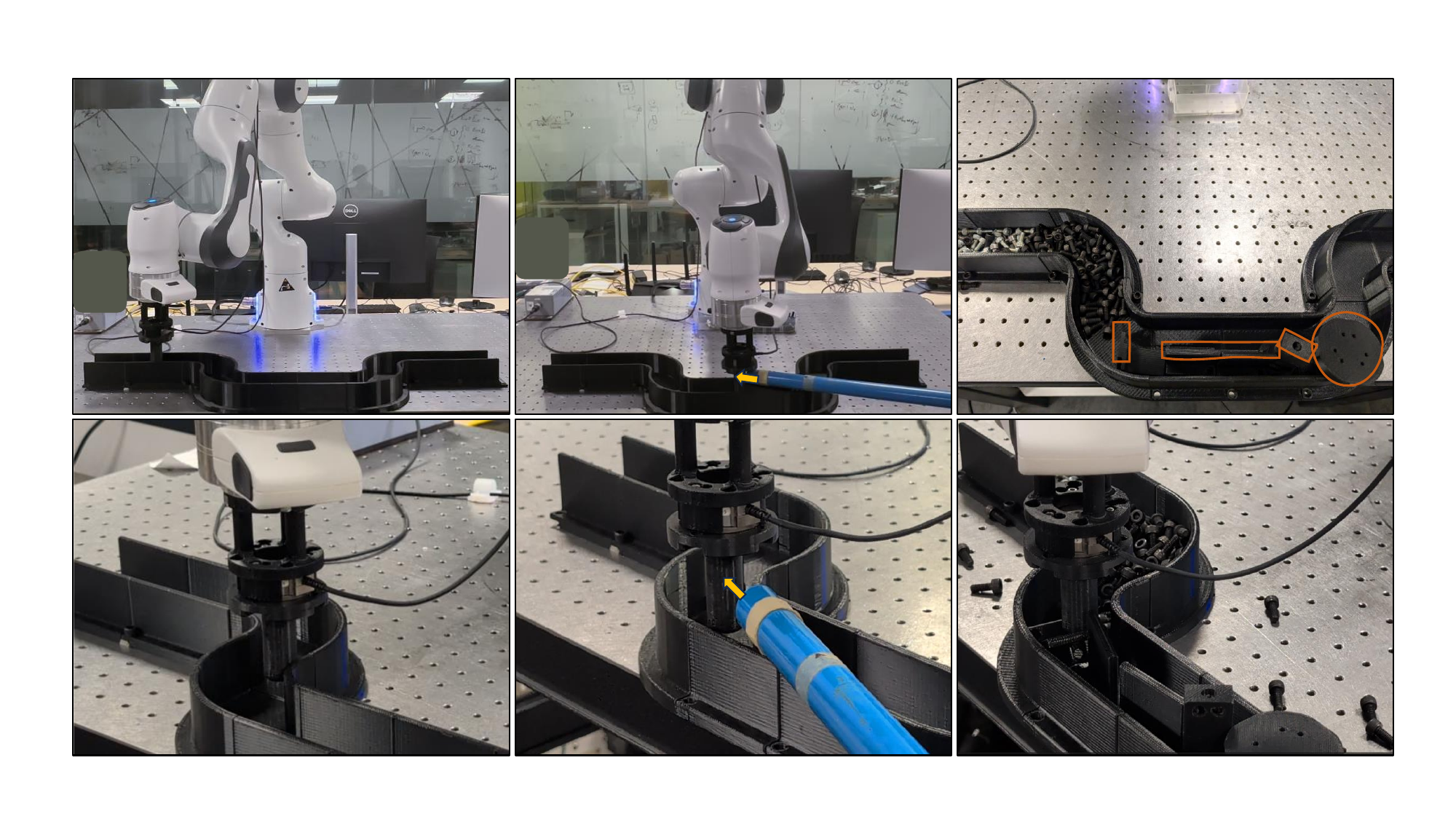}
    \\
    \vspace{3mm}
    \includegraphics[width=0.99\linewidth]{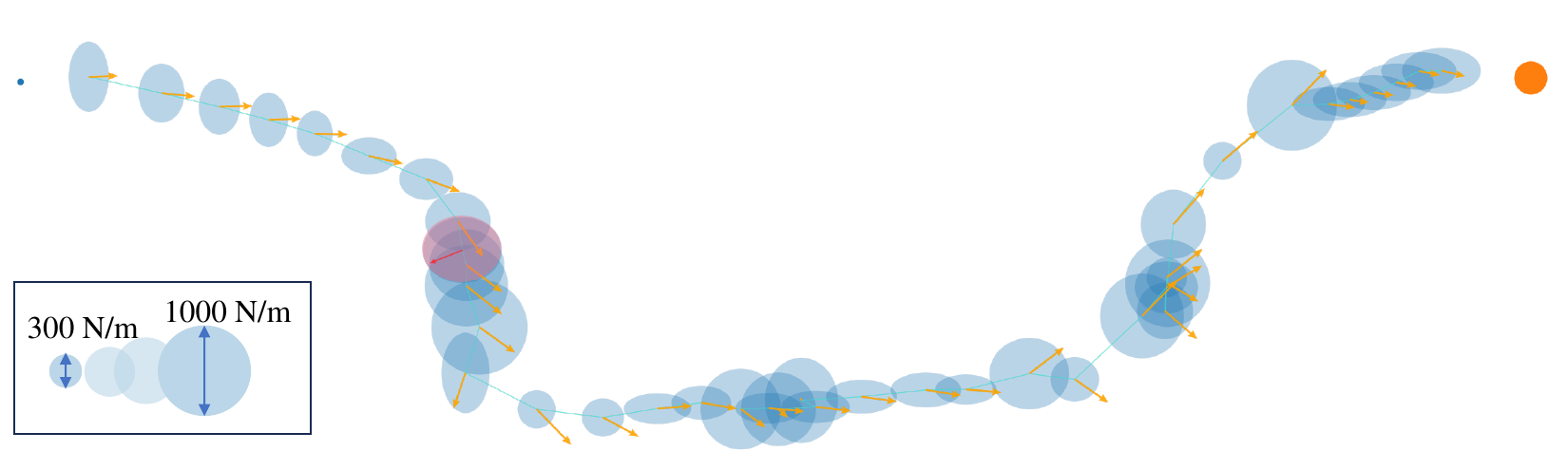}
    \caption{Real-world setup. Top: Overview of the entire setup. Bottom: Closeups of the robot in action. Left: The robot makes contact with a wall. Center: A stick applies random external forces during motion. Right: The robot interacts with large, stationary obstacles.
    Bottom: the trajectory of the external-force run shows robot behavior, where the ellipses indicate the stiffness. 
    The red ellipse indicates the recovery action. 
    }
    \label{fig:real_world}
\end{figure}

\addtolength{\textheight}{-1.5cm}
\subsection{Real-world experiments}\label{sec:real_world_exp}
We deployed the policy on a physical 7-DOF Franka robot arm without any fine-tuning. In the setup, the RL policy is run on a separate ROS
node that communicates with the physical robot running in a real-time frequency $1000 Hz$. 
We trained the best performing seed from Sec.~\ref{sec:maze-task} further in 1500 episodes instead of 500. 
We tested the performance with 10 consecutive runs in a comprehensive set of scenarios, including added obstacles and human perturbation. The results are presented in Table~\ref{tab:real_world}
and Fig.~\ref{fig:real_world}.
Detailed video is available at the~\href{https://jack-sherman01.github.io/Bresa}{project website}. We present a highly authentic video without cuts or speedup adjustments, showcasing 10 consecutive cycles of task execution from two perspectives (overall and closeup of end-effector) simultaneously.

\section{Discussion and Limitation} 
In this section, we discuss the limitations of our method.
The offline data collection is a crucial step in our method, as it determines the performance of \textit{safety critic}. However, the offline data collection process requires a large number of samples to ensure the safety critic's accuracy. 
Furthermore, the performance of our method is highly dependent on the hyperparameters, such as the action size and discount factors. The optimal hyperparameters may vary depending on the task and environment, making it challenging to find the best configuration. In the future, we plan to explore automated hyperparameter tuning methods to optimize the performance of our method.
Lastly, although our method shows promising results in both simulation and real-world experiments, it is only validated in one single contact-rich task. In the future, we plan to adopt more complex tasks with more challenging environments to showcase the performance of our method.

\section{Conclusion and future work}
We presented Bresa, a novel hierarchical reinforcement learning method designed to enhance safety in contact-rich robotic tasks. Inspired by biological reflexes, our approach decouples task learning and safety learning, allowing a risk critic to operate at a higher frequency than the task-solving policy. By integrating low-level risk-aware control with variable impedance control (VIC), our method ensures real-time intervention in unsafe situations while maintaining adaptability in dynamic and unstructured environments. Experimental results demonstrate that our method outperforms baselines, improving real-time safety during physical interactions. While Bresa enhances safety in contact-rich tasks, several directions remain open for future exploration, such as multi-modal safety mechanisms (vision or tactile sensing), which will strengthen the reflexes capability.   


\bibliographystyle{IEEEtran}
\bibliography{root}

\begin{thebibliography}{10}
\providecommand{\url}[1]{#1}
\csname url@rmstyle\endcsname
\providecommand{\newblock}{\relax}
\providecommand{\bibinfo}[2]{#2}
\providecommand\BIBentrySTDinterwordspacing{\spaceskip=0pt\relax}
\providecommand\BIBentryALTinterwordstretchfactor{4}
\providecommand\BIBentryALTinterwordspacing{\spaceskip=\fontdimen2\font plus
\BIBentryALTinterwordstretchfactor\fontdimen3\font minus \fontdimen4\font\relax}
\providecommand\BIBforeignlanguage[2]{{%
\expandafter\ifx\csname l@#1\endcsname\relax
\typeout{** WARNING: IEEEtran.bst: No hyphenation pattern has been}%
\typeout{** loaded for the language `#1'. Using the pattern for}%
\typeout{** the default language instead.}%
\else
\language=\csname l@#1\endcsname
\fi
#2}}

\bibitem{suomalainen2022survey}
M.~Suomalainen, Y.~Karayiannidis, and V.~Kyrki, ``A survey of robot manipulation in contact,'' \emph{Robotics and Autonomous Systems}, vol. 156, p. 104224, 2022.

\bibitem{gu2024safesurvey}
S.~Gu, L.~Yang, Y.~Du, G.~Chen, F.~Walter, J.~Wang, and A.~Knoll, ``A review of safe reinforcement learning: Methods, theory and applications,'' \emph{IEEE Transactions on Pattern Analysis and Machine Intelligence}, 2024.

\bibitem{dietterich2000hierarchical}
``Hierarchical reinforcement learning with the maxq value function decomposition,'' \emph{Journal of artificial intelligence research}, vol.~13, pp. 227--303, 2000.

\bibitem{martin2019variable}
R.~Mart{i}n-Mart{i}n, M.~A. Lee, R.~Gardner, S.~Savarese, J.~Bohg, and A.~Garg, ``Variable impedance control in end-effector space: An action space for reinforcement learning in contact-rich tasks,'' in \emph{2019 IEEE/RSJ International Conference on Intelligent Robots and Systems (IROS)}.\hskip 1em plus 0.5em minus 0.4em\relax IEEE, 2019, pp. 1010--1017.

\bibitem{brunke2022safe}
L.~Brunke, M.~Greeff, A.~W. Hall, Z.~Yuan, S.~Zhou, J.~Panerati, and A.~P. Schoellig, ``Safe learning in robotics: From learning-based control to safe reinforcement learning,'' \emph{Annual Review of Control, Robotics, and Autonomous Systems}, vol.~5, pp. 411--444, 2022.

\bibitem{bharadhwaj2020conservative}
H.~Bharadhwaj, A.~Kumar, N.~Rhinehart, S.~Levine, F.~Shkurti, and A.~Garg, ``Conservative safety critics for exploration,'' \emph{arXiv preprint arXiv:2010.14497}, 2020.

\bibitem{thananjeyan2021recovery}
B.~Thananjeyan, A.~Balakrishna, S.~Nair, M.~Luo, K.~Srinivasan, M.~Hwang, J.~E. Gonzalez, J.~Ibarz, C.~Finn, and K.~Goldberg, ``Recovery rl: Safe reinforcement learning with learned recovery zones,'' \emph{IEEE Robotics and Automation Letters}, vol.~6, no.~3, pp. 4915--4922, 2021.

\bibitem{HengSRL-VIC}
H.~Zhang, G.~Solak, G.~J.~G. Lahr, and A.~Ajoudani, ``Srl-vic: A variable stiffness-based safe reinforcement learning for contact-rich robotic tasks,'' \emph{IEEE Robotics and Automation Letters}, vol.~9, no.~6, pp. 5631--5638, 2024.

\bibitem{nguyen2021robust}
Q.~Nguyen and K.~Sreenath, ``Robust safety-critical control for dynamic robotics,'' \emph{IEEE Transactions on Automatic Control}, vol.~67, no.~3, pp. 1073--1088, 2021.

\bibitem{noseworthy2025forge}
M.~Noseworthy, B.~Tang, B.~Wen, A.~Handa, C.~Kessens, N.~Roy, D.~Fox, F.~Ramos, Y.~Narang, and I.~Akinola, ``Forge: Force-guided exploration for robust contact-rich manipulation under uncertainty,'' \emph{IEEE Robotics and Automation Letters}, 2025.

\bibitem{ajoudani2018progress}
A.~Ajoudani, A.~M. Zanchettin, S.~Ivaldi, A.~Albu-Sch{\"a}ffer, K.~Kosuge, and O.~Khatib, ``Progress and prospects of the human-robot collaboration,'' \emph{Autonomous Robots}, vol.~42, no.~5, pp. 957--975, 2018.

\bibitem{kuo2021uncertainty}
C.-Y. Kuo, A.~Schaarschmidt, Y.~Cui, T.~Asfour, and T.~Matsubara, ``Uncertainty-aware contact-safe model-based reinforcement learning,'' \emph{IEEE Robotics and Automation Letters}, vol.~6, no.~2, pp. 3918--3925, 2021.

\bibitem{bear2020neuroscience}
M.~Bear, B.~Connors, and M.~A. Paradiso, \emph{Neuroscience: Exploring the brain, enhanced edition: Exploring the brain}.\hskip 1em plus 0.5em minus 0.4em\relax Jones \& Bartlett Learning, 2020.

\bibitem{fan2024learn}
K.~Fan, Z.~Chen, G.~Ferrigno, and E.~De~Momi, ``Learn from safe experience: Safe reinforcement learning for task automation of surgical robot,'' \emph{IEEE Transactions on Artificial Intelligence}, vol.~5, no.~7, pp. 3374--3383, 2024.

\bibitem{liu2023safe}
P.~Liu, K.~Zhang, D.~Tateo, S.~Jauhri, Z.~Hu, J.~Peters, and G.~Chalvatzaki, ``Safe reinforcement learning of dynamic high-dimensional robotic tasks: navigation, manipulation, interaction,'' in \emph{2023 IEEE International Conference on Robotics and Automation (ICRA)}.\hskip 1em plus 0.5em minus 0.4em\relax IEEE, 2023, pp. 9449--9456.

\bibitem{bing2023safety}
Z.~Bing, A.~Mavrichev, S.~Shen, X.~Yao, K.~Chen, K.~Huang, and A.~Knoll, ``Safety guaranteed manipulation based on reinforcement learning planner and model predictive control actor,'' \emph{arXiv preprint arXiv:2304.09119}, 2023.

\bibitem{zhang2023efficient}
X.~Zhang, C.~Wang, L.~Sun, Z.~Wu, X.~Zhu, and M.~Tomizuka, ``Efficient sim-to-real transfer of contact-rich manipulation skills with online admittance residual learning,'' in \emph{Conference on Robot Learning}.\hskip 1em plus 0.5em minus 0.4em\relax PMLR, 2023, pp. 1621--1639.

\bibitem{aflakian2024robust}
A.~Aflakian, J.~Hathaway, R.~Stolkin, and A.~Rastegarpanah, ``Robust contact-rich task learning with reinforcement learning and curriculum-based domain randomization,'' \emph{IEEE Access}, 2024.

\bibitem{zhu2022contact}
X.~Zhu, S.~Kang, and J.~Chen, ``A contact-safe reinforcement learning framework for contact-rich robot manipulation,'' in \emph{2022 IEEE/RSJ International Conference on Intelligent Robots and Systems (IROS)}.\hskip 1em plus 0.5em minus 0.4em\relax IEEE, 2022, pp. 2476--2482.

\bibitem{altman1995constrained}
E.~Altman, \emph{Constrained Markov decision processes}.\hskip 1em plus 0.5em minus 0.4em\relax Routledge, 1995.

\bibitem{ott2008cartesian}
C.~Ott, \emph{Cartesian impedance control of redundant and flexible-joint robots}.\hskip 1em plus 0.5em minus 0.4em\relax Springer, 2008.

\bibitem{SAC}
\BIBentryALTinterwordspacing
T.~Haarnoja, A.~Zhou, P.~Abbeel, and S.~Levine, ``Soft actor-critic: Off-policy maximum entropy deep reinforcement learning with a stochastic actor,'' in \emph{Proceedings of the 35th International Conference on Machine Learning}, ser. Proceedings of Machine Learning Research, J.~Dy and A.~Krause, Eds., vol.~80.\hskip 1em plus 0.5em minus 0.4em\relax PMLR, 10--15 Jul 2018, pp. 1861--1870. [Online]. Available: \url{https://proceedings.mlr.press/v80/haarnoja18b.html}
\BIBentrySTDinterwordspacing

\bibitem{lillicrap2015continuous}
T.~P. Lillicrap, J.~J. Hunt, A.~Pritzel, N.~Heess, T.~Erez, Y.~Tassa, D.~Silver, and D.~Wierstra, ``Continuous control with deep reinforcement learning,'' \emph{arXiv preprint arXiv:1509.02971}, 2015.

\end{thebibliography}

\end{document}